# Symmetry as a Representation of Intuitive Geometry?

**Wangcheng Xu (wxu324@gmail.com)**

**Snejana Shegheva (shegheva@gmail.com)**

**Ashok K. Goel (ashok.goel@cc.gatech.edu)**

Design & Intelligence Laboratory, School of Interactive Computing, Georgia Institute of Technology,
Atlanta, USA

## Abstract

Recognition of geometrical patterns seems to be an important aspect of human intelligence. Geometric pattern recognition is used in many intelligence tests, including Dehaene's odd-one-out test of Core Geometry (CG)) based on intuitive geometrical concepts (Dehaene et al., 2006). Earlier work has developed a symmetry-based cognitive model of Dehaene's test and demonstrated performance comparable to that of humans. In this work, we further investigate the role of symmetry in geometrical intuition and build a cognitive model for the 2-Alternative Forced Choice (2-AFC) variation of the CG test (Marupudi & Varma 2021). In contrast to Dehaene's test, 2-AFC leaves almost no space for cognitive models based on generalization over multiple examples. Our symmetry-based model achieves an accuracy comparable to the human average on the 2-AFC test and appears to capture an essential part of intuitive geometry.

## Introduction

Geometric intelligence refers to recognition, reasoning and learning of geometric concepts, relations, and patterns. Research in cognitive science recently has witnessed much work on geometric intelligence in large part because of the presence of geometric problems on tests of general human intelligence (Bringsjord 2011; Carpenter, Just & Shell 1990; Dastani & Indurkhya 2001; Evans 1969; Kunda, McGreggor & Goel 2013; Lovett & Forbus 2011; Lovett, Lockwood & Forbus 2008; McGreggor, Kunda & Goel 2014; Santoro et al. 2017, 2018; Schwering et al. 2007; Shegheva & Goel 2018; Zhang et al. 2019a, 2019b). The study of geometric intelligence by constructing computational models of intelligence tests provides essential insights into human cognition that complement the results of other experiments.

Figure 1(a) illustrates an example[1] from the Standard Raven's Progressive Matrices (RPM) test of intelligence (Raven, Raven & Court 1998), and Figure 1(b) illustrates an example from the Core Geometry (CG) test (Dehaene et al. 2006). RPM is a very common test used around the world to measure general human intelligence. In RPM, given a 3x3 matrix of geometric figures with one entry at the bottom right missing (top of Figure 1(a)), the task is to select a figure from a set of eight choices (bottom of Figure 1(a)) that would best complete the pattern in the matrix. Problems on RPM are classified into multiple categories that test different types of geometric inferences. In CG, given a set of six geometric figures (Figure 1(b)), the task is to select the odd one out. As with RPM, problems on CG are classified into multiple categories that test different geometric abilities such as Euclidean geometry, metric properties, chirality, etc.

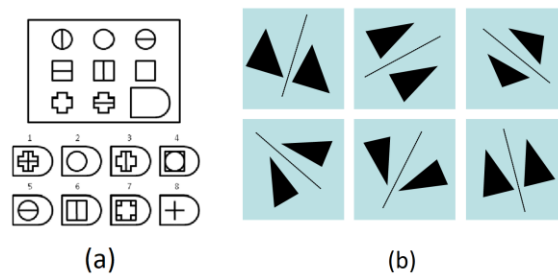

Figure 1(a) and 1(b): An illustrative problem from the Raven's Progressive Matrices test (on the left) and the Core Geometry test (on the right).

A major question in cognitive science is what part of human geometric knowledge (if any) is "innate" in that it developed over biological evolution, or "intuitive" in that it is part of unconscious cognition (Kahneman 2011; Stanovich & West 2000). Dehaene et al. (2006) suggest that geometric knowledge needed to address the CG test may be intuitive (or perhaps innate). When they administered the CG test to humans schooled in the modern educational system and to unschooled tribal people, they found that the tribal people performed about as well on the CG test as people who had had the benefit of formal schooling. Their results have been replicated by similar studies such as Izard et al. (2011).

Various computational models of geometric intelligence make different assumptions about the intuitiveness of geometric knowledge. For example, the Lovett and colleagues' model of RPM (Lovett, Lockwood & Forbus 2008) and CG (Lovett & Forbus 2011) is based on analogy with the basic set of geometric figures: it assumes prior knowledge of geometric concepts such as triangles and closed figures, and analogical generalizations across a set of figures often are derived at problem-solving time. In contrast,

---

[1] Due to copyright reasons, we have included an example similar to the RPM test rather than an actual example from the test.

Shegheva & Goel's model of RPM (Shegheva & Goel 2018) and CG (Shegheva & Goel 2021) is based on the use of symmetry; it assumes prior knowledge mainly of symmetry.

Recently, Marupudi & Varma (2020, 2021) have proposed a 2 Alternative Forced Choice variation (2-AFC) of the CG test. As Figure 2 illustrates, Marupudi & Varma's 2-AFC variation consists of selecting one out of two figures that are most similar to a given figure. Problems on their 2-AFC test cover the same range of geometric inferences that CG covers. Interestingly, and perhaps also a little surprisingly, Marupudi & Varma (2021) found that humans perform much better than chance at the 2-AFC test. As they point out, this argues against a model based on induction because the top row in 2-AFC consists of only one figure.

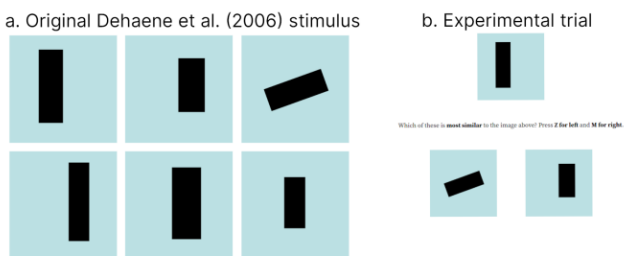

Figure 2: An illustrative example of the 2-AFC problem adapted from Marupudi & Varma (2021). The original stimulus by Dehaene et al. (2006) (on the left) and the corresponding experimental 2-AFC trial (on the right).

This helps frame the research question in our work: can the symmetry method address the 2-AFC test, and, if so, how well does it relate to human performance on the test?

## Related Work

The literature on addressing geometry problems on intelligence tests is rapidly growing. Carpenter, Just & Shell (1990) describe two early models of the Raven's Progressive Matrices Test of general human intelligence. Their models express geometrical patterns in the form of production rules. Earlier, Evans (1969) had built a (partial) computational model that used analogy to address problems similar to (though simpler than) the RPM test. Forbus and colleagues have long been strong proponents of analogy as the core process for addressing geometry problems on intelligence tests. They have used the structure-mapping theory of analogy as the basis for developing computational models for the RPM test (Lovett, Lockwood & Forbus 2008) as well as the Dehaene's Core Geometry test (Lovett & Forbus 2011). The performance of their models for these tasks is comparable to that of humans. Bringsjord (2011) has proposed the performance in intelligent tests as a measure of progress in AI.

The literature on the role of symmetry in visual perception is large. In both computer vision and computer graphics, symmetry enables analysis of geometrical shapes in terms of their mathematical properties (Liu et al. 2010). In compelling research on human perception described in Generative Theory of Shape, Leyton (2001) emphasizes the role of symmetry in understanding complex shapes constructed from primitive forms via deconstructive representations.

Symmetry transformations lie at the core of Gestalt principles of visual perception (Wagemans et al. 2012; Wertheimer 1945). Dastani & Indurkhya (2001) and Schwering et al. (2007) have used Gestalt principles to understand (simple) geometric proportional analogies. In observing a set of shapes related via direct or latent features, symmetry provides a framework for studying the generative processes of the shapes as well as their relationships with one another figures. Shegheva & Goel (2021) demonstrate the usefulness of applying Gestalt principles to Dehaene's CG images for identifying features that highlight potential "symmetry-breaking". In a related computational model addressing the RPM test, Shegheva & Goel (2018) developed a method that used structure alignment to detect patterns of relationships between images at a pixel level. The two models achieved performance comparable to human performance on the RPM and CG tests, respectively.

Earlier, Kunda, McGreggor & Goel (2013) applied affine transformations such as translation, rotation, and reflection (typical symmetry operations) directly to the pixel-level representations of the images in the RPM test. This research was based on evidence that human perception applies similar operations to visual images (Kosslyn, Thompson & Ganis 2006). The results showed a significant correlation with the visual strategies used by individuals with autistic traits (Kunda & Goel 2011). McGreggor, Kunda & Goel (2014) proposed a fractal representation that captures self-similarity of an image. Their technique executes an automatic adjustment of the level that allows viewing of the images under different magnifications to assess their similarity. They successfully used the fractal approach for various visual tasks such as RPM, CG, and other Odd-One-Out tests, and demonstrated a performance comparable to that of humans (McGreggor & Goel 2013). Common to all these computational models is the idea that symmetry captures a strictly intuitive sense of visual perception in which shapes are analyzed under mental transformations.

Some recent work has explored the use of convolutional neural networks and deep learning to address geometric problems similar to those on the RPM test (Santoro et al. 2017, 2018; Zhang et al. 2019b). However, this line of work typically requires the construction of large synthetic data sets of RPM-like problems (Zhang et al. 2019a) that are implausible from the perspective of human learning. In any case, this research line has not yet addressed the CG test and thus is not applicable to 2-AFC problems deriving from CG.

## 2-AFC Testing Trials

Like Marupudi & Varma's construction of experimental 2-AFC problems (2021), our 2-AFC testing trials are generated directly from Dehaene's CG test and cover 43 geometric concepts in 7 categories. As shown in Figure 2, each 2-AFC trial includes the only image not following the concept as one choice and two images that embody the concept as the target image and the other choice, respectively. Figure 3 illustrates example trials for different concepts from each of the seven categories: topology, Euclidean geometry, geometrical figures, symmetrical figures, chiral figures, metric properties, and geometrical transformations.

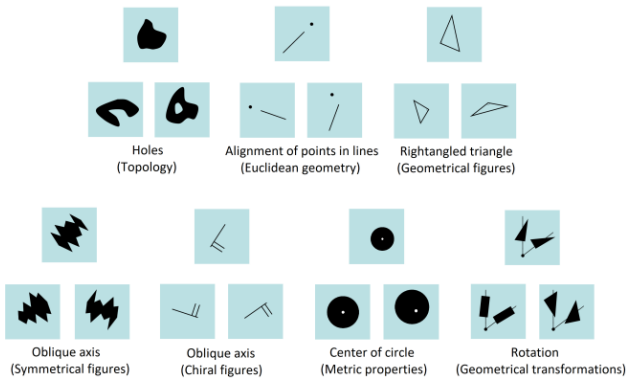

Figure 3: Seven examples of the 2-AFC trial, one from each category. Each trial includes three images. The target image is at the top, and two choices are shown at the bottom.

As shown in Figure 4, for our experiments, we built twenty 2-AFC trials for every CG concept, including all possible permutations from every CG problem. This allows an accuracy measurement at a granularity of 5% for each concept.

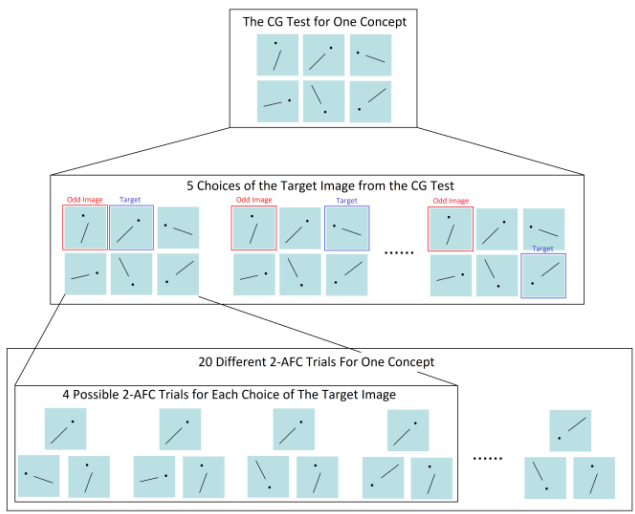

Figure 4: Generation of all possible 2-AFC trials for one concept from the original stimulus by Dehaene et al. (2006).

## A Symmetry-Based Cognitive Model of 2-AFC Problems

### Overall Structure

As in Shegheva & Goel's model for Dehaene's CG test (2021), the core of our symmetry-based model for 2-AFC problems is the symmetry transformation of images through Principal Component Analysis (PCA). However, our model also considers symmetry at the feature level by introducing self-symmetry after the symmetry-based alignment via PCA. To conserve chirality, our computational model uses only the first principal component of PCA, and the images are aligned with the first principal axis through rotation. As a result, two alternative orientations are possible, and both need to be considered as indicated in Figure 5.

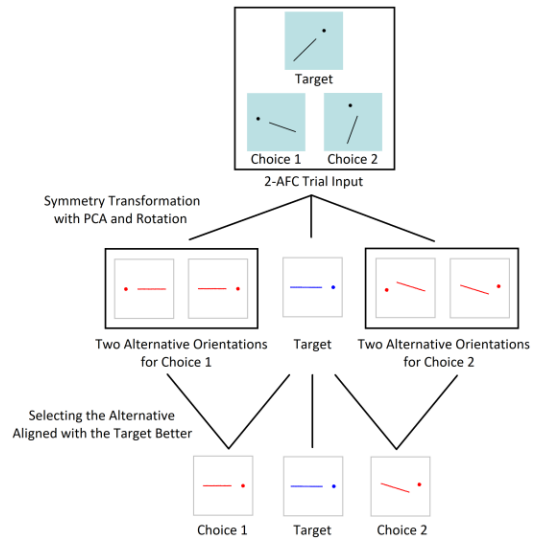

Figure 5: Symmetry transformation for a 2-AFC trial. For each choice, the alternative with less difference from the target out of the two possible orientations is selected.

Measurement of dissimilarity between images is based on a set of features. Our model looks at three simple features (center-shift, area, and spread) that capture the basic characteristics of the geometrical figures in the image space and the self-symmetry of these base features. The decision on the image orientations too is based on the measurements with the same three base features. The subsequent image comparisons during answer selection consider a preset subset of all six features, including the base features and self-symmetry features, with equal weights. For a pair of images, the overall difference is computed by adding the differences in the chosen features. The choice image with a less overall difference from the target image is selected as the answer.

### Image Preprocessing

For the convenience of computation in later stages, all input images are converted into binary ones based on a preset threshold before any processing. The binary image distinguishes the pixels belonging to the geometrical figure from those belonging to the background.

**Symmetry Transformation**

Visual artifacts for measuring intelligence may contain various noise sources - measurement errors such as image alignment issues and intentional transformations applied to test the brain's agility in the presence of irrelevant features. As Figure 5 illustrates, the first step in our symmetry-based model involves the symmetry transformation of images that addresses alignment, position, and orientation noise while accenting the features relevant to the geometry concepts in the given task.

For each image, the geometrical figure's pixels are extracted as a set of 2D points, with each point as the horizontal and vertical coordinates of the corresponding pixel. Applying PCA to the set of points finds the first principal axis of the figure. Then, the original binary image is centered and aligned with the first principal axis through rotation.

The rotation may align the image in two different orientations, which are the 180-degree rotated images of each other. Our model checks both orientations of each choice image and chooses the one with less a difference from the target image using the base features, ensuring the ideal alignment with the target image after the transformations. The three transformed images, like the ones shown at the bottom in Figure 5, are used for further comparison.

**Feature Extraction**

Understanding geometrical concepts may require multiple priors ranging from simple geometrical tasks of scaling and rotating the objects to extracting topological features indicating object containment inside or outside of a perimeter. This paper aims to implement a model that is limited to a set of only symmetry priors and evaluate its ability to perform on the extended range of concepts as described in the 2-AFC problems.

To test the dissimilarity between images, our model considers three base symmetry features that represent the basic statistical characteristics: center shift, area, and spread. While one feature may correlate with other features on certain geometrical concepts, each feature reflects different aspects of the image space.

The computation of features is based on the set of points A that belong to the geometrical figure in the transformed images. In all equations below, X[a] and Y[a] denote the x and y coordinates of point a, respectively.

**Center Shift** The center shift measures how the center of slices of the figure shifts along the axis. Thus, it detects the symmetry of the figure along the axis. The center shift C along the vertical and horizontal dimensions, $C_v$ and $C_h$, are:
- $C_v(y) = \frac{1}{k}\sum_{i=1}^{k} X[a_i]$, for k points $\in \{a_i | Y[a_i] = y\}$
- $C_h(x) = \frac{1}{k}\sum_{i=1}^{k} Y[a_i]$, for k points $\in \{a_i | X[a_i] = x\}$

**Area** The area measures the number of pixels in the slices of the figure along the axis. It captures the distribution of the figure's mass in that dimension. The area A along the vertical and horizontal dimensions, $A_v$ and $A_h$, are as follows:
- $A_v(y) = \sum_{i=1}^{k} 1$, for k points $\in \{a_i | Y[a_i] = y\}$
- $A_h(x) = \sum_{i=1}^{k} 1$, for k points $\in \{a_i | X[a_i] = x\}$

**Spread** The spread measures how the standard deviation of slices of the figure changes along the axis. It may correlate with the area in many figures but can capture shapes that expand outward against those with similar areas but concentrate around the center. The spread S along the vertical and horizontal dimensions, $S_v$ and $S_h$, are as follows:
- $S_v(y) = \sqrt{\frac{1}{k}\sum_{i=1}^{k}\left(X[a_i] - \frac{1}{k}\sum_{i=1}^{k} X[a_i]\right)^2}$,
for k points $\in \{a_i | Y[a_i] = y\}$
- $S_h(x) = \sqrt{\frac{1}{k}\sum_{i=1}^{k}\left(Y[a_i] - \frac{1}{k}\sum_{i=1}^{k} Y[a_i]\right)^2}$,
for k points $\in \{a_i | X[a_i] = x\}$

**Reasoning with Base Features**

The absolute differences in the base features between the choice and target images capture their dissimilarity. Figure 6 shows an example of feature extraction and the calculation of the difference in the horizontal dimension. Our model also performs a similar extraction and computation in the vertical dimension. For each base feature, the difference is the sum in the vertical and horizontal dimensions.

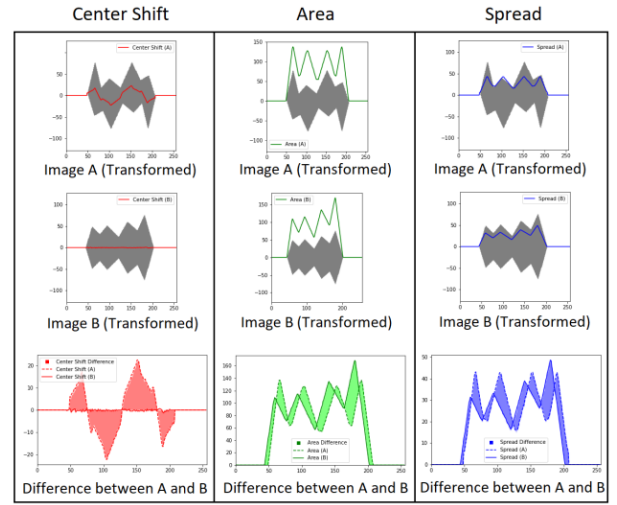

Figure 6: An example of three features in the horizontal dimension. For each feature, the absolute difference between the two images is visualized with overlayed curves in the bottom row. The scale may vary for each plot. The center shift shows the most difference, which captures the dissimilarity in the embodied geometrical concept.

**Reasoning with Self-Symmetry**

Reasoning over geometrical concepts requires selecting from a range of symmetry operations – reflection, rotation,

translation, and scaling. Our base features, along with the PCA transformation, account for variances of the same geometrical objects induced by rotation or translation. However, noisy attributes such as size variations or hollow vs. solid textures may overwhelm the base features due to their scale and intensity and weaken the geometrical concept's signal. To re-balance observed features and amplify anomalies corresponding to the concept, we add self-symmetry to the feature extraction phase that examines differences between the object's vertical and horizontal dimensions. Figure 7 illustrates a concept of a circle's center that is relatively easy for a human to detect. Unlike the base feature, self-symmetry can capture the off-center shift by suppressing the size difference between figures and amplifying the differences between dimensions within an image. Self-symmetry highlights attributes that violate rotation invariant properties of symmetry operations.

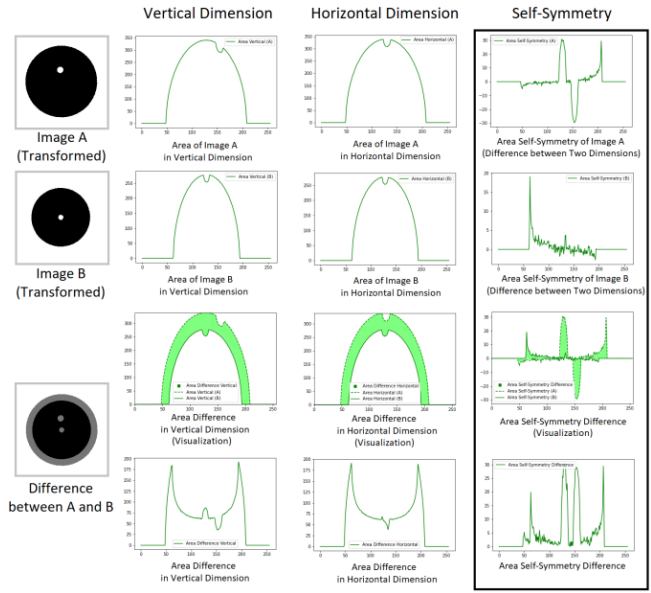

Figure 7: An example of the self-symmetry of the area feature. In the first two rows, the difference between the first two columns gives the self-symmetry. The third row in all three columns is the visualization that overlays the first two rows and highlights the difference. The fourth row is the corresponding curve for the difference. The scale may vary for each plot. Self-symmetry captures the off-center shift as two spikes in the middle of the bottom right plot.

### Choice Selection

For each choice image, our model sums its differences from the target image for a selection of the base and self-symmetry features. The model selects the choice with a less overall difference as the answer.

### Results

The results for one base feature and the best combinations with two and four features in all seven categories are summarized in Table 1. The four-feature mixed combination achieves the highest overall accuracy at 84.7%, slightly better than the average human accuracy. The mixed combination with both base features and self-symmetry features tends to have better performance since the two variants complement each other. The two-feature mixed combination outperforms any subset of only base features or self-symmetry ones.

Table 1: The overall category-wise accuracy for the top base feature and the best combinations of two and four features.

| Model Accuracy by Category | | | |
|---|---|---|---|
| Category | Center Shift (Base) | Center Shift (Base), Spread (Self-Symmetry) | Center Shift (Base), Center Shift, Area, Spread (Self-Symmetry) |
| Symmetrical figures | 90% | 85% | 87% |
| Chiral figures | 100% | 100% | 100% |
| Euclidean geometry | 88% | 87% | 87% |
| Geometrical figures | 63% | 83% | 86% |
| Geometrical transformations | 73% | 75% | 78% |
| Metric properties | 90% | 90% | 91% |
| Topology | 55% | 58% | 64% |
| Overall Accuracy | 78.4% | 82.8% | 84.7% |

The category-wise and concept-wise comparisons between the model's performance and the average human accuracy adapted from Marupudi & Varma's experiment (2021) are shown in Table 2 and Table 3, respectively. The concept-wise comparisons (Table 3) only include the four-feature model.

Table 2: The category-wise differences and standard deviations (STD) between the model and human accuracy that are sorted ascendingly by the four-feature model's STD. The difference data bar is centered at zero, and a positive difference means higher accuracy for the model. Human data is adapted from Marupudi & Varma (2021).

| Comparison of Model Accuracy with Human Accuracy by Category | | | | | | | |
|---|---|---|---|---|---|---|---|
| Category | Human Accuracy | Center Shift (Base) | | Center Shift (Base), Spread (Self-Symmetry) | | Center Shift (Base), Center Shift, Area, Spread (Self-Symmetry) | |
| | | Difference | STD | Difference | STD | Difference | STD |
| Symmetrical figures | 82.0% | 8.0% | 8.7% | 3.0% | 9.9% | 4.7% | 11.2% |
| Chiral figures | 96.2% | 18.0% | 14.1% | 18.0% | 14.1% | 18.0% | 14.1% |
| Euclidean geometry | 91.2% | -8.0% | 12.9% | -9.3% | 14.4% | -9.3% | 14.8% |
| Geometrical figures | 71.4% | -28.4% | 27.2% | -7.9% | 17.5% | -5.7% | 14.9% |
| Geometrical transformations | 76.3% | 1.1% | 16.9% | 3.6% | 16.1% | 6.8% | 16.0% |
| Metric properties | 82.0% | 13.7% | 16.8% | 13.7% | 17.9% | 14.4% | 16.7% |
| Topology | 82.2% | -27.2% | 22.2% | -24.7% | 28.2% | -18.5% | 26.7% |
| Overall Difference/STD | | -5.3% | 20.0% | -0.9% | 18.4% | 1.0% | 17.3% |

The significantly better-than-chance accuracy for our symmetry-based model on the 2-AFC test indicates that it captures elements of intuitive geometry. However, the performance varies from category to category, surpassing average human accuracy by a considerable margin in chiral figures and metric properties while falling behind in topology. In general, our model performs better for the geometrical concepts with stronger visual similarity. For instance, the geometrical figures for metric properties share similar shapes and sizes even if arranged in random orientations. The same applies to the chiral figures.

Conversely, our model has the lowest accuracy in topology since the figures may have arbitrarily different shapes.

Overall, the comparison of the performance of our four-feature symmetry-based model with human performance shown in Table 2, seems to indicate fair match for six of the seven categories in the 2-AFC test (symmetrical figures, chiral figures, Euclidean geometry, geometrical figures, geometrical transformations, and metric properties), with the seventh category of topology being the exception. Table 3 provides a more detailed comparison of our model's accuracy and human accuracy on each problem within the different categories.

Table 3: The concept accuracy difference between the best four-feature model and the human average. Categories are ordered by ascending STD, with concepts within a category ordered by descending difference. The difference data bar is produced similarly to that in Table 2. The trial concept numbers are the same as in Deheane et al. (2006) and the human data are adapted from Marupudi & Varma (2021).

| Concept Number | Concept Category | Geometrical Concepts | Human Accuracy | Accuracy Difference |
|---|---|---|---|---|
| 28 | Symmetrical figures | Vertical axis | 85.8% | 14.2% |
| 30 | | Oblique axis | 87.5% | 2.5% |
| 29 | | Horizontal axis | 72.7% | -2.7% |
| 44 | Chiral figures | Oblique axis | 65.5% | 34.5% |
| 38 | | Oblique axis | 71.6% | 28.4% |
| 42 | | Vertical axis | 94.3% | 5.7% |
| 41 | | Vertical axis | 96.6% | 3.4% |
| 14 | Euclidean geometry | Right angle | 94.3% | 5.7% |
| 11 | | Alignment of points in lines | 94.9% | 5.1% |
| 10 | | Straight line | 96.0% | 4.0% |
| 15 | | Right angle | 97.7% | 2.3% |
| 7 | | Alignment of points in lines | 98.3% | 1.7% |
| 8 | | Curve | 96.0% | -11.0% |
| 40 | | Secant lines | 93.2% | -38.2% |
| 37 | | Parallel lines | 98.9% | -43.9% |
| 23 | Geometrical figures | Square | 85.2% | 14.8% |
| 9 | | Convex shape | 93.8% | 6.3% |
| 26 | | Trapezoid | 89.2% | 5.8% |
| 17 | | Circle | 96.6% | 3.4% |
| 20 | | Equilateral triangle | 97.2% | 2.8% |
| 24 | | Rectangle | 94.3% | -4.3% |
| 25 | | Parallelogram | 88.6% | -18.6% |
| 12 | | Quadrilateral | 94.9% | -19.9% |
| 13 | | Rightangled triangle | 81.3% | -41.3% |
| 33 | Geometrical transformations | Horizontal symmetry | 62.5% | 37.5% |
| 34 | | Rotation | 48.3% | 36.7% |
| 35 | | Oblique symmetry | 76.7% | 23.3% |
| 36 | | Homothecy (fixed orientation) | 74.4% | 10.6% |
| 39 | | Homothecy (fixed size) | 71.0% | 9.0% |
| 27 | | Vertical symmetry | 73.1% | -3.1% |
| 31 | | Translation | 81.3% | -16.3% |
| 32 | | Point symmetry | 83.5% | -43.5% |
| 22 | Metric properties | Center of quadilateral | 47.7% | 37.3% |
| 19 | | Middle of segment | 68.2% | 31.8% |
| 45 | | Increasing distance | 74.4% | 25.6% |
| 21 | | Fixed proportion | 72.7% | 17.3% |
| 18 | | Center of circle | 90.3% | 9.7% |
| 16 | | Distance | 96.6% | 3.4% |
| 43 | | Equidistance | 84.1% | -24.1% |
| 6 | Topology | Connectedness | 81.3% | 18.8% |
| 5 | | Closure | 81.3% | 3.8% |
| 4 | | Inside | 97.7% | -47.7% |
| 3 | | Holes | 68.8% | -48.8% |

## Discussion and Conclusions

Recently research in cognitive science has advanced several cognitive models of geometric intelligence, especially for geometric problems manifested on tests of general human intelligence. The various models make different assumptions about prior knowledge. Studies in cognitive science, such as Dehaene et al. 2006, have conjectured that some geometric knowledge may be innate (the result of biological evolution) or intuitive (part of unconscious automatic cognition). However, it has been unclear what this exactly means for various models of geometric intelligence. Marupudi & Varma's 2-AFC test of geometric intelligence provides a clearer test. As they have argued, it appears to rule out induction or generalization as a plausible model because the test offers only one input figure and generalization requires multiple figures.

In this paper, we described the use of a symmetry model for the 2-AFC test of geometric intelligence. Earlier, Sheghava & Goel had shown the usefulness of the symmetry model for addressing problems on the RPM and the CG tests. We found that the overall performance of the symmetry model on the 2-AFC test of geometric intelligence is comparable to that of humans for six of the seven categories of problems in Marupudi & Varma's study.

Measured through the limited set of 2-AFC trials generated from Dehaene's CG test, our symmetry-based model with the best feature combination achieves a slightly higher overall accuracy than the average human. At the same time, there are a few noticeable concepts where the symmetry model has far lower accuracy, such as the right-angled triangle, point symmetry, and two topological concepts.

Our symmetry model raises a few interesting and complementary hypotheses for further exploration. First, human cognition in general may use multiple representations and reasoning strategies, including both symmetry and generalization. For the RPM and CG tests, both reasoning strategies may be available and applicable. Second, while problems on the RPM and CG tests may superficially appear to be instances of visual analogies, symmetry may offer a deeper explanation as it better generalizes to the 2-AFC test. Third, symmetry might be part of the geometric knowledge that Dehaene et al. (2006) consider being intuitive. Symmetry has the advantage of making minimal assumptions about prior knowledge as it assumes no knowledge about geometric concepts such as a circle or parallel lines: symmetry itself is the intuitive knowledge.

Future work on the symmetry model itself may explore multiple issues. The realization that symmetry captures intuitive information about geometrical shapes and their transformations allows for formulating representations that can apply to various visual perception tasks and possibly generalizing them into more complex shapes and relationships. In addition, the critical aspect of realigning images using the PCA technique provides a foundation for an attention mechanism that reduces ambiguity in the presence of noise features, such as scaling, random axes, and spatial positions. Additional research on the nuances of the PCA-based algorithm to self-adjust to the type and the complexity of the problem can benefit in multiple directions, for example, pattern detection in a sequence of images, reconstruction of geometrical operations, or geometry concept enhancement through irrelevant feature reduction.

## Acknowledgments
Much of this work was done when all three authors were with Georgia Tech's Design & Intelligence Laboratory; Wangcheng Xu is now a Ph.D. student at Northwestern University. We are very grateful to Sashank Varma and Vijay Marupudi for sharing their work, results, and unpublished manuscript with us. This work has also benefited from many discussions with Keith McGreggor.